\documentclass{article}
\usepackage{spconf,amsmath,graphicx,hyperref}

\title{ConCAD: Constraint-Aware Image-to-CAD Generation with Dual-Granularity Rewards}
\name{\begin{tabular}{c}
Chenxi Zhai \qquad Xi Cheng \qquad Hang Cheng \qquad Zhicheng Guan \\
Mingyu Fan \qquad Yanzhe Tang \qquad Pingfa Feng \qquad Long Zeng$^*$\thanks{$^*$Corresponding author.}
\end{tabular}}

\address{Tsinghua University, Shenzhen International Graduate School }

\begin{document}
\ninept
\maketitle
\begin{abstract}

Image-to-CAD generation seeks executable parametric programs that recover both the geometry and design intent of a reference object. Existing systems are commonly evaluated by validity and shape overlap, although two solids with similar volume can encode different CAD relations. We introduce ConCAD, a constraint-aware image-to-CAD framework optimized via Group Relative Policy Optimization (GRPO) with rewards at two complementary granularities: a code-level constraint reward  and an execution-level geometric reward. This complementary design disambiguates structurally distinct yet volumetrically similar shapes while ensuring valid 3D geometry. To verify that these rewards recover geometry and design intent, we introduce a B-rep geometric constraint satisfaction rate (G-CSR), which analytically extracts and evaluates geometric constraints from boundary representations. Experiments on the DeepCAD and Zero2CAD demonstrate that ConCAD achieves the best IoU and Chamfer Distance over competitive baselines, while also outperforming them on G-CSR, validating its superior recovery of both geometric fidelity and parametric design intent.
\end{abstract}

%图像到CAD的生成旨在寻找能够恢复参考对象的几何形状和设计意图的可执行参数化程序。现有系统通常通过有效性和形状重叠度进行评估，但体积相近的两个实体可能编码不同的CAD关系。为此，我们提出了 ConCAD——一种通过群组相对策略优化（GRPO）进行训练的约束感知图像到CAD生成框架，并在两个互补粒度上构建奖励机制：代码级约束奖励和执行级几何奖励。这种互补设计能够消解“体积相似但结构不同”的形状歧义，同时确保生成有效的 3D 几何实体。为了验证这些奖励机制能够真正恢复几何形状与设计意图，我们提出了基于 B-rep 的几何约束满足率（G-CSR），从边界表示中解析提取并评估几何约束。在 DeepCAD 和 Zero2CAD上的实验表明，ConCAD 相比竞争基线取得了最优的 IoU 和倒角距离（Chamfer Distance），同时在 G-CSR 指标上也全面优于现有方法，充分证明了其在几何保真度与参数化设计意图恢复上的双重优势。
%
\section{Introduction}
Reconstructing parametric CAD programs from images underpins key applications in reverse engineering \cite{varady1997reverse}, editable asset creation \cite{willis2021fusion}, and mechanical design automation. Recent vision-language models (VLMs) translate rendered parts directly into CAD programs~\cite{doris2026cad,alrashedy2025generating,guan2026cad,wang2025cad,li2025cadllamaleveraginglargelanguage,li2024cad}, bypassing task-specific decoders while inheriting rich program priors. Nevertheless, existing systems are predominantly evaluated by execution validity, Chamfer Distance (CD) \cite{yang2018foldingnet}, and volumetric Intersection-over-Union (IoU) \cite{doris2026cadbench}. While these metrics indicate whether solids look alike, they fail to assess whether the generated design preserves the underlying relations that make CAD models structured and editable.

Design intent is intrinsically expressed through geometric constraints such as coaxiality, symmetry, and concentricity. Crucially, such relations are easily violated even when global IoU remains high. Evaluating intent purely through source-code text similarity is inadequate, as geometrically identical solids can stem from vastly different scripts. Nonetheless, extracting structural relations directly from code offers distinct advantages during learning: it delivers dense feedback on partially correct or non-executable programs by capturing syntactic cues and operational structures before the policy reliably yields valid boundary representations.

We therefore decouple policy optimization from intent verification and propose ConCAD. Trained via GRPO \cite{shao2024deepseekmath}, ConCAD incorporates dual-granularity supervision coupling the code-level constraint reward ($R_{\text{con}}$) with the execution-level geometry reward ($R_{\text{iou}}$). Specifically, the code-level constraint reward ($R_{\text{con}}$) provides dense structural feedback prior to kernel execution, while the execution-level geometry reward ($R_{\text{iou}}$) enforces spatial fidelity through inertia-normalized volumetric overlap. For rigorous evaluation, we introduce the B-rep Geometric Constraint Satisfaction Rate (G-CSR) to analytically quantify topological relations directly on solid boundaries. Our contributions are threefold:

\begin{itemize}
    \item We formulate image-to-CAD synthesis as constraint-aware parametric program generation, presenting the ConCAD framework that optimizes policy training via GRPO to align editable programs with visual inputs.
    \item We propose a dual-granularity reward scheme integrating the code-level constraint reward with the execution-level geometry reward, simultaneously enforcing topological constraint consistency and 3D spatial fidelity.
    \item We introduce G-CSR to quantify relational design constraints directly on B-rep topology, and demonstrate across the DeepCAD and Zero2CAD datasets that ConCAD achieves state-of-the-art geometric fidelity and topological constraint preservation.
\end{itemize}

% 从图像中重构参数化 CAD 程序，是逆向工程、可编辑资产生成以及机械设计自动化等核心应用的关键支撑。近期的视觉语言模型（VLM）能够直接将渲染零件转化为 CadQuery 代码，不仅绕过了特定任务的序列解码器，还继承了丰富的程序生成先验。然而，现有系统大多以执行有效性、倒角距离（CD）或体素交并比（IoU）为评估核心。这些指标只能回答两个实体在外观上是否相似，却无法评估生成的模型是否保留了赋予 CAD 模型结构化与可编辑性的内在关系。

% 设计意图本质上是通过几何实体间的约束来表达的：孔特征可能共轴或对称，轮廓线可能同心或相切，而平面或草图线段可能平行、正交或呈等尺寸阵列。至关重要的是，即便全局 IoU 很高，这类关系依然可能被打破。仅凭文本层面的源代码相似度来评估设计意图并不理想，因为几何形状相同的实体可以通过截然不同的 CadQuery 脚本构建出来。尽管如此，在学习过程中直接从代码中提取结构性设计关系具有显著优势：即便模型尚未能可靠地生成有效的 B-rep（边界表示）模型，这种方法也能通过捕捉代码中保留的语法特征、操作结构及参数比例，为部分正确甚至无法执行的程序提供密集的奖励反馈。

% 为此，我们提出了 ConCAD（约束感知图像到CAD生成框架）。在群组相对策略优化（GRPO）过程中，ConCAD 将代码级约束奖励与执行级几何奖励相结合。具体而言，代码级奖励（R_con）提取结构化设计关系进行反馈，而几何奖励（R_iou）则通过惯量归一化的体积重叠度约束三维形状。在评测上，我们引入了 B-rep 几何约束满足率（G-CSR），直接在实体边界上对拓扑关系进行解析。

% 本文的贡献总结如下：
    % 1. 我们将 image-to-CAD 合成形式化为约束感知的参数化程序生成，提出了通过 GRPO 优化策略训练的 ConCAD 框架，以使可编辑程序与视觉输入精准对齐。
    % 2. 我们提出了一个双粒度奖励机制，将 the code-level constraint reward 与 the execution-level geometry reward 相结合，同时确保拓扑约束一致性与 3D 空间保真度。
    % 3. 我们提出了 G-CSR 指标，直接在 B-rep 拓扑上量化关联设计约束，并在 DeepCAD 与 Zero2CAD 数据集上证明了 ConCAD 在几何保真度与拓扑约束保持两方面均达到了领先性能。
\section{Related Work}
\noindent\textbf{Parametric CAD generation.} 
Early CAD generation approaches represent the modeling process as structured command sequences. DeepCAD~\cite{wu2021deepcad} encodes and decodes CAD sequence tokens via dedicated autoregressive networks. Text2CAD~\cite{khan2024text2cad}, CAD-recode~\cite{rukhovich2025cad}, CAD-Diffuser~\cite{ma2024draw}, Img2CAD~\cite{li2026image2cadseq}, and CADCrafter~\cite{chen2025cadcrafter} introduce text, point cloud, or image conditioning, employing autoregressive models, diffusion frameworks, or conditional factorizations to bolster cross-modal synthesis. More recent approaches shift toward executable code generation, leveraging the rich programming priors of vision-language models (VLMs) to produce editable scripts, as exemplified by CAD-Coder~\cite{doris2026cad} and Zero2CAD~\cite{ataei2026zero}. While framing CAD synthesis as code generation inherits powerful priors, standard supervised training inherently relies on surface-level token imitation, struggling to capture parametric dependencies and underlying geometric constraints.

\noindent\textbf{Reinforcement learning for CAD.} 
To enhance code executability and spatial reasoning, recent efforts incorporate reinforcement learning (RL) into CAD program generation. Cadrille~\cite{kolodiazhnyi2025cadrille} introduces an online reinforcement learning strategy to interactively refine program synthesis. CADCoder~\cite{guan2026cad}combines CoT prompting with RL fine-tuning. ReCAD~\cite{li2026recad} optimizes generation policy by coupling aligned volumetric IoU with rendered visual feature similarity as rewards. Despite their effectiveness, these methods rely predominantly on execution-level geometric surrogates or rendered visual cues, leaving the optimization process agnostic to whether underlying parametric design relations are faithfully satisfied.

% 参数化 CAD 生成：早期的 CAD 生成方法将建模过程表示为结构化的命令序列，DeepCAD 通过专用编码器和自回归解码器对命令序列进行建模。Text2CAD、CAD-Diffuser、Img2CAD 和 CADCrafter 进一步引入文本、点云或图像等模态条件，采用自回归、扩散模型或条件分解等机制来提升跨模态生成能力。近期的方法则将 CAD 建模视为可执行代码生成任务，利用视觉语言模型（VLM）丰富的编程先验来生成可编辑脚本，如 CAD-Coder 与 Zero2CAD。然而，尽管代码化表征继承了强大的语言先验，标准的监督微调本质上仍依赖词元级的表层模仿，难以捕捉参数化设计依赖与底层几何约束。

% CAD 的强化学习：为了提升代码的可执行性与空间推理能力，近期的工作开始将思维链（CoT）推理与强化学习（RL）引入 CAD 生成领域。例如，CADCoder 结合了 CoT 提示机制与强化学习微调，以强化程序合成与几何推理过程；ReCAD 则通过融合对齐体积 IoU 与视觉特征相似度来构建复合奖励。尽管这些方法取得了显著效果，但它们主要依赖执行级的几何代理指标或渲染视觉线索，优化过程依然缺乏对参数化设计关系是否真正满足的显式约束。

\begin{figure*}[t] % 跨栏通常只支持放在页面顶部 [t]
    \centering
    % 宽度使用 \textwidth（整页文本宽度），而不是 \columnwidth（单栏宽度）
    \includegraphics[width=0.8\textwidth]{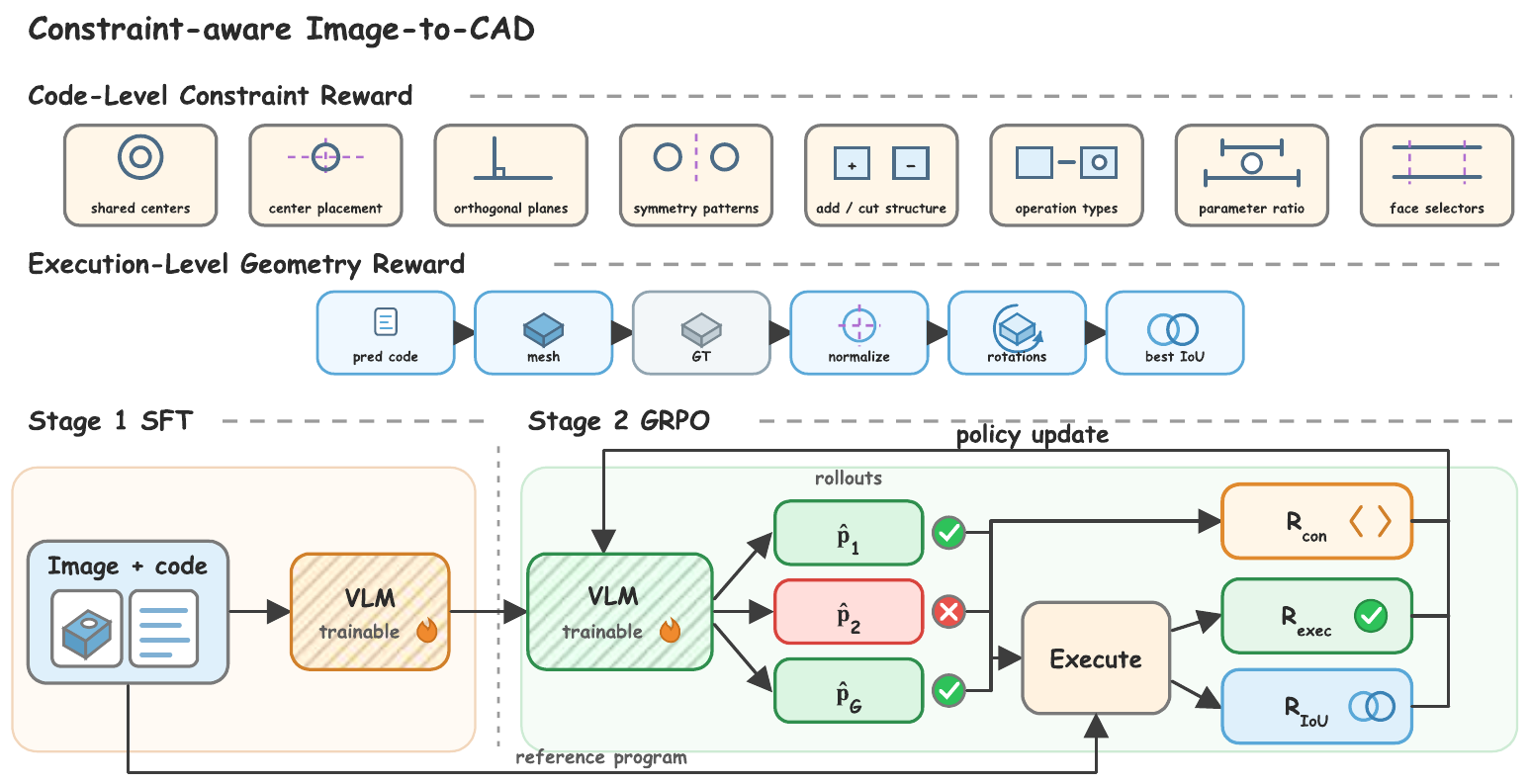} 
    \caption{Overview of the ConCAD training pipeline. Following supervised initialization, the policy samples $G$ candidate programs from an input image and is optimized via Group Relative Policy Optimization (GRPO), combining the pre-execution code-level constraint reward $R_{\rm con}$ with the execution-level geometry reward $R_{\rm iou}$ computed via inertia-aligned volumetric overlap on valid solids. This complementary formulation ensures both parametric design intent and 3D spatial fidelity.}
    \label{fig:framework}
\end{figure*}
\vspace{-3pt}
\section{Method}

\subsection{Problem Formulation}
Given a single rendered image $I$, our goal is to synthesize an executable CadQuery program $\hat{p} \sim \pi_\theta(\cdot \mid I)$ that instantiates an editable 3D solid $\hat{\Omega} = \mathcal{E}(\hat{p})$ through a CAD modeling kernel $\mathcal{E}$. Let $p^*$ denote the ground-truth program and $\Omega^* = \mathcal{E}(p^*)$ its reference boundary representation (B-rep). While standard supervised fine-tuning (SFT) optimizes policy $\pi_\theta$ via next-token prediction, it struggles to capture structural design intent and execution robustness. We therefore formulate image-to-CAD generation as a reinforcement learning problem, seeking to maximize the expected reward over rollouts:
\begin{equation}
\max_\theta \, \mathbf{E}_{I \sim \mathcal{D}, \, \hat{p} \sim \pi_\theta(\cdot \mid I)} [R(I, \hat{p}, p^*)].
\end{equation}
The overall training pipeline of ConCAD with dual-granularity rewards is illustrated in Fig.~\ref{fig:framework}.

% \section{方法}

% \subsection{问题定义}
% 给定单张渲染图像 $I$，目标是合成一段可执行的 CadQuery 程序 $\hat{p} \sim \pi_\theta(\cdot \mid I)$，通过 CAD 内核 $\mathcal{E}$ 执行后实例化为一个可编辑的三维实体 $\hat{\Omega} = \mathcal{E}(\hat{p})$。设 $p^*$ 为真实参考程序，$\Omega^* = \mathcal{E}(p^*)$ 为其对应的实体边界表示（B-rep）。传统的监督微调（SFT）通过最大化词元似然来训练策略 $\pi_\theta$，而后续的强化学习方法大多仅针对执行成功的样本优化几何层面的相似度 $d(\hat{\Omega}, \Omega^*)$。

% 然而，仅依赖几何相似度忽略了参数化设计意图，同时，对源代码 $p^*$ 进行简单的文本匹配也无法识别几何功能等价的代码。为此，ConCAD 构建了一套双粒度奖励优化机制，将代码级约束奖励 $R_{\text{con}}$ 与执行级几何对齐奖励 $R_{\text{iou}}$ 有机结合，并由执行有效性门控 $R_{\text{exec}}$ 予以保障。在训练完成后的评测阶段，我们引入了基于 B-rep 的几何约束满足率指标（$\mathrm{G\mbox{-}CSR}$），在实体拓扑层面检查模型的参数化设计关系。

\subsection{Policy Optimization via GRPO}
We initialize $\pi_\theta$ via supervised fine-tuning on paired data, setting $\pi_{\text{ref}} = \pi_\theta$. In the RL stage, Group Relative Policy Optimization (GRPO)~\cite{shao2024deepseekmath} optimizes $\pi_\theta$ without a critic network. For each input image $I$, the policy samples rollouts $\{\hat{p}_i\}_{i=1}^{G} \sim \pi_\theta(\cdot \mid I)$, each scored by:
\begin{equation}
    R_i = \lambda_{\text{e}} R_{\text{exec}, i} + \lambda_{\text{c}} R_{\text{con}, i} + \lambda_{\text{i}} R_{\text{iou}, i},
\end{equation}
with balancing weights empirically set to $(0.1, 0.3, 1.0)$. Within-group advantages are standardized as $A_i = (R_i - \mu_R) / (\sigma_R + \epsilon)$, where $\mu_R$ and $\sigma_R$ denote the group statistics. The policy is optimized via the clipped surrogate objective with a KL penalty:
\begin{equation}
    \mathcal{L}(\theta) = -\frac{1}{G}\sum_{i=1}^G \mathcal{L}_{\text{clip}}(\hat{p}_i, A_i; \theta) + \beta D_{\text{KL}}(\pi_\theta \parallel \pi_{\text{ref}}),
\end{equation}
where $\mathcal{L}_{\text{clip}}$ denotes the standard PPO clipping term. This relative ranking assigns differential advantages to unexecutable rollouts preserving valid parametric patterns, mitigating exploration plateaus caused by sparse geometric feedback.

% \subsection{基于 GRPO 的策略优化}
% 我们首先在图像-程序配对数据上通过监督微调（SFT）训练策略网络作为参考模型 $\pi_{\text{ref}}$。在强化学习阶段，采用群组相对策略优化（GRPO）~\cite{grpo} 更新 $\pi_\theta$，无需训练辅助价值网络。

% 对于输入图像 $I$，策略网络采样一组候选程序 $\{\hat{p}_i\}_{i=1}^{G} \sim \pi_\theta(\cdot \mid I)$。每个样本 $\hat{p}_i$ 分配一个复合标量奖励：
% \begin{equation}
%     R_i = \lambda_{\text{exec}} R_{\text{exec}, i} + \lambda_{\text{con}} R_{\text{con}, i} + \lambda_{\text{iou}} R_{\text{iou}, i},
% \end{equation}
% 其中各奖励项的具体形式见第 \ref{sec:rewards} 节，超参数设为 $(\lambda_{\text{exec}}, \lambda_{\text{con}}, \lambda_{\text{iou}}) = (0.10, 0.30, 1.00)$。组内优势值标准化计算为：
% \begin{equation}
%     A_i = \frac{R_i - \mu_R}{\sigma_R + \epsilon}, \quad \mu_R = \frac{1}{G}\sum_{i=1}^G R_i, \quad \sigma_R = \sqrt{\frac{1}{G}\sum_{i=1}^G (R_i - \mu_R)^2}。
% \end{equation}
% 策略网络通过最大化带有参考模型 KL 散度约束的截断代理目标进行更新：
% \begin{equation}
%     \mathcal{L}_{\text{GRPO}}(\theta) = -\frac{1}{G}\sum_{i=1}^G \min \left( \frac{\pi_\theta(\hat{p}_i \mid I)}{\pi_{\theta_{\text{old}}}(\hat{p}_i \mid I)} A_i, \, \operatorname{clip}\left(\frac{\pi_\theta(\hat{p}_i \mid I)}{\pi_{\theta_{\text{old}}}(\hat{p}_i \mid I)}, 1-\varepsilon, 1+\varepsilon\right) A_i \right) + \beta \mathbb{D}_{\text{KL}}(\pi_\theta \parallel \pi_{\text{ref}})。
% \end{equation}
% 组内相对排序机制确保了具有部分正确代码结构的探索样本依然能获得正向梯度反馈。

\subsection{Dual-Granularity Reward Formulation}
Relying on a naive execution signal easily induces reward hacking, driving the policy toward trivial, easily executable primitives (e.g., degenerating into a simple cube); conversely, without the grounding of execution, optimizing code-level rewards in isolation tends to produce syntactically dense yet unexecutable scripts. We therefore design a dual-granularity reward scheme, where the execution validity gate serves as a foundational baseline, while code-level constraints and execution-level geometric signals provide complementary guidance.

\noindent\textbf{Foundational Execution Validity Gate ($R_{\text{exec}}$).} 
Because generated code may harbor syntax errors, infinite loops, or topological anomalies, all candidate programs are executed within isolated subprocesses guarded by strict timeouts. The CAD executor $\mathcal{E}$ parses the program, constructs the underlying OpenCASCADE topology, and exports a triangle mesh. We assess program viability via a gating signal $R_{\text{exec}} \in \{0, 1\}$:
\begin{equation}
    R_{\text{exec}} = \mathbf{1}\left[\mathcal{E}(\hat{p}) \text{ succeeds} \ \wedge \ \mathrm{Vol}(\hat{\Omega}) > 0\right].
\end{equation}
If execution crashes, times out, or produces a degenerate solid, $R_{\text{exec}} = 0$.

\noindent\textbf{Granularity 1: Code-Level Constraint Reward ($R_{\mathrm{con}}$).} 
A rule-based pattern matcher inspects the syntactic patterns and operational structures of $\hat{p}$ against $p^*$ via regular expressions without CAD kernel execution. The matcher summarizes design intent across an eight-relation engineering constraint set $\mathcal{K}_{\mathrm{code}}$, including geometric datums and positional constraints, modeling features and operation distributions, numerical parameters and proportions, as well as mirror and pattern configurations. For each constraint category $k \in \mathcal{K}_{\mathrm{code}}$, let $n_k^*$ denote the number of ground-truth relations in $p^*$, and $s_k$ the number of satisfied relations matched in $\hat{p}$. The code-level constraint reward is computed as the average ratio:
\begin{equation}
R_{\mathrm{con}} = \frac{\sum_{k \in \mathcal{K}_{\mathrm{code}}} s_k}{\sum_{k \in \mathcal{K}_{\mathrm{code}}} n_k^*}.
\label{eq:rcon}
\end{equation}
This reward provides effective feedback for generated programs possessing valid structural patterns.

\noindent\textbf{Granularity 2: Execution-Level Geometric Reward ($R_{\text{iou}}$).} 
When $R_{\text{exec}} = 1$, we evaluate the physical geometric agreement between $\hat{\Omega}$ and $\Omega^*$. To resolve scale ambiguity in single-view images as well as coordinate frame variations arising from different CadQuery modeling workflows, each solid is first centered and normalized by its inertia-derived characteristic radius:
\begin{equation}
    n(\Omega) = \left\{ \frac{\mathbf{x} - \bar{\mathbf{x}}}{\sqrt{\operatorname{tr}(\mathbf{I}) / (2\operatorname{Vol}(\Omega))}} : \mathbf{x} \in \Omega \right\},
\end{equation}
where $\bar{\mathbf{x}}$ is the center of mass and $\mathbf{I}$ is the moment-of-inertia tensor. To resolve sign ambiguities in principal axes, let $U_{\hat{\Omega}}$ and $U_{\Omega^*}$ denote the principal-inertia eigenbases of the predicted and reference solids, respectively. We compute candidate alignment rotations using the four proper sign-flip matrices:
\begin{equation}
\begin{aligned}
    Q_s &= U_{\Omega^*} S_s U_{\hat{\Omega}}^{\mathsf{T}}, \quad S_s = \operatorname{diag}(\boldsymbol{\sigma}_s), \\
    \text{s.t.} \quad & \boldsymbol{\sigma}_s \in \{\pm 1\}^3, \ \det S_s = 1, \ s \in \{1, 2, 3, 4\}.
\end{aligned}
\end{equation}
The continuous geometric alignment reward $R_{\text{iou}}$ is defined as the maximum volumetric intersection-over-union across identity and the sign-flip rotations $\mathcal{Q} = \{I\} \cup \{Q_s\}_{s=1}^4$:
\begin{equation}
    R_{\text{iou}} = \max_{Q \in \mathcal{Q}} \operatorname{IoU}\left(n(\Omega^*), Q \cdot n(\hat{\Omega})\right).
\end{equation}
In practice, volumetric IoU is estimated via Monte Carlo sampling with 8,192 points in the joint bounding container. When $R_{\text{exec}} = 0$, we set $R_{\text{iou}} = 0$.

\subsection{Evaluation Metric: G-CSR}
To evaluate the capability of generative models in recovering underlying parametric design intent, we establish the Boundary Representation (B-rep) Geometric Constraint Satisfaction Rate (G-CSR), which is specifically tailored to inspect authentic relational geometric constraints between solid entities (e.g., coaxiality, concentricity, tangency, parallelism, perpendicularity, and symmetry).

After normalizing scales and aligning orientations via 
$\mathcal{Q}$, greedy bipartite matching pairs predicted and reference B-rep primitives under predefined tolerances to verify topological relations.Let $\mathcal{K}$ denote the set of active constraint categories; for category $k \in \mathcal{K}$, with $n_k^*$ reference constraints and $s_k$ satisfied constraints, $\mathrm{G\mbox{-}CSR}$ is formulated as the macro-averaged satisfaction rate:
\begin{equation}
    \mathrm{G\mbox{-}CSR} = \frac{1}{|\mathcal{K}|} \sum_{k \in \mathcal{K}} \frac{s_k}{n_k^*}.
\end{equation}
This metric quantifies the parametric topological consistency of 3D solids, effectively compensating for the limitation of superficial geometric metrics in capturing engineering design logic.

% \subsection{评测指标：G-CSR}
% 为评估生成模型对底层参数化设计意图的恢复能力，我们构建了基于边界表示（B-rep）的几何约束满足率指标（$\mathrm{G\mbox{-}CSR}$）。该指标专用于检验实体间真实的几何关系约束（如共轴、同心、相切、平行、垂直及对称等），不参与策略训练。

% 我们首先从执行成功的实体中提取解析图元（面与边），并通过局部尺寸归一化与候选旋转集 $\mathcal{Q}$ 消除尺度和位姿歧义。随后，在设定几何公差下利用贪心二分图匹配建立预测与真实图元的一对一映射，在此基础上逐项检验实体拓扑关系。设有效约束类别集合为 $\mathcal{K}$，对于类别 $k \in \mathcal{K}$，参考约束数为 $n_k^*$，预测实体中被满足的约束数为 $s_k$，$\mathrm{G\mbox{-}CSR}$ 定义为各几何约束类别的宏观平均满足率：
% \begin{equation}
%     \mathrm{G\mbox{-}CSR} = \frac{1}{|\mathcal{K}|} \sum_{k \in \mathcal{K}} \frac{s_k}{n_k^*}.
% \end{equation}
% 该指标直接量化了三维实体的参数化拓扑一致性，有效弥补了表层几何指标（如 IoU）无法反映工程设计逻辑的缺陷。

\section{Experiments}
\subsection{Experimental Setup}
\noindent\textbf{Datasets.} 
We evaluate on two CAD datasets: DeepCAD~\cite{wu2021deepcad} and Zero2CAD~\cite{ataei2026zero}. For DeepCAD, we adopt the dataset processed by CAD-Coder~\cite{doris2026cad}, comprising 147k training instances and the official test split. For Zero2CAD, we use the Zero2CAD-100K dataset for training and validation, restricting inputs strictly to a single-view setting for fair alignment.

\noindent\textbf{Evaluation Protocol.} 
To ensure strict comparability, our evaluation adheres to the benchmark protocol established by CAD-Coder~\cite{doris2026cad}, reporting four complementary metrics: 
Invalidity Ratio (IR, percentage of unexecutable or empty outputs), 
IoU (computed via exact boolean intersection of solids normalized and aligned via principal inertia axes under proper sign-flips), 
Chamfer Distance (CD), and 
G-CSR measuring topological constraint satisfaction on successfully executed solids.

\noindent\textbf{Implementation Details.} 
We adopt Qwen3-VL-2B-Instruct\cite{yang2025qwen3} as our common backbone and train all models on a single NVIDIA A800-80GB GPU. We first perform SFT for one epoch over the full dataset with an effective batch size of 8 and an initial learning rate of $10^{-5}$ to yield the reference policy $\pi_{\text{ref}}$. GRPO fine-tuning is scheduled for 800 updates with $G=4$ rollouts per prompt, a sampling temperature of 0.8. On DeepCAD, GRPO samples from a constraint-rich subset filtered to prevent reward saturation on trivial primitives.  Both SFT and GRPO stages are optimized using AdamW paired with a cosine annealing learning rate schedule.

% \section{实验}
% \subsection{实验设置}
% \noindent\textbf{数据集。} 
% 我们在两个 CAD 基准上展开评估：\textbf{DeepCAD} 与 \textbf{Zero2CAD}。对于 DeepCAD，我们采用经由 CAD-Coder 处理构建的基准数据，该数据将 DeepCAD 几何模型转化为可执行的 CadQuery 程序并配对多视角渲染图，涵盖 147k 训练样本并提供了官方的测试集。对于 Zero2CAD，我们使用 Zero2CAD-100K 数据集进行训练和验证，并在全流程中严格限制为单视角输入以确保公平对齐。

% \noindent\textbf{评测协议。} 
% 为确保严格可比性，我们的测试参照 CAD-Coder 的评测规范，并报告四项互补指标：
% 无效率（\textbf{IR}$\downarrow$：执行失败或空实体的占比）、
% 对齐体积交并比（\textbf{IoU}$\uparrow$：基于主惯性轴中心化、归一化及符号翻转对齐后精确布尔相交体积比）、
% 倒角距离（\textbf{CD}$\downarrow$）、以及
% 用于衡量成功执行实体拓扑约束满足率的 \textbf{G-CSR}$\uparrow$。

% \noindent\textbf{实现细节。} 
% 我们采用 Qwen3-VL-2B-Instruct 作为通用骨干模型，并使用单张 NVIDIA A800-80GB GPU 训练。首先在完整数据集上执行一个周期的全参数 SFT，有效批次大小为 8，初始学习率为 $10^{-5}$，由此获得参考策略 $\pi_{\text{ref}}$。随后 GRPO 强化学习设置 800 步更新，每张图像采样 $G=4$ 条路径，采样温度为 0.8，有效批次大小为 4，初始学习率为 $10^{-6}$，KL 散度惩罚系数设为 $\beta=0.001$。在 DeepCAD 上，GRPO 从包含圆、圆弧或多次拉伸特征的强约束子集中采样，避免模型在简单基础图元上发生奖励饱和。SFT 与 GRPO 阶段均采用 AdamW 优化器并结合余弦退火学习率调度策略。

\begin{figure*}[!t]
\centering
    \includegraphics[width=0.6\textwidth]{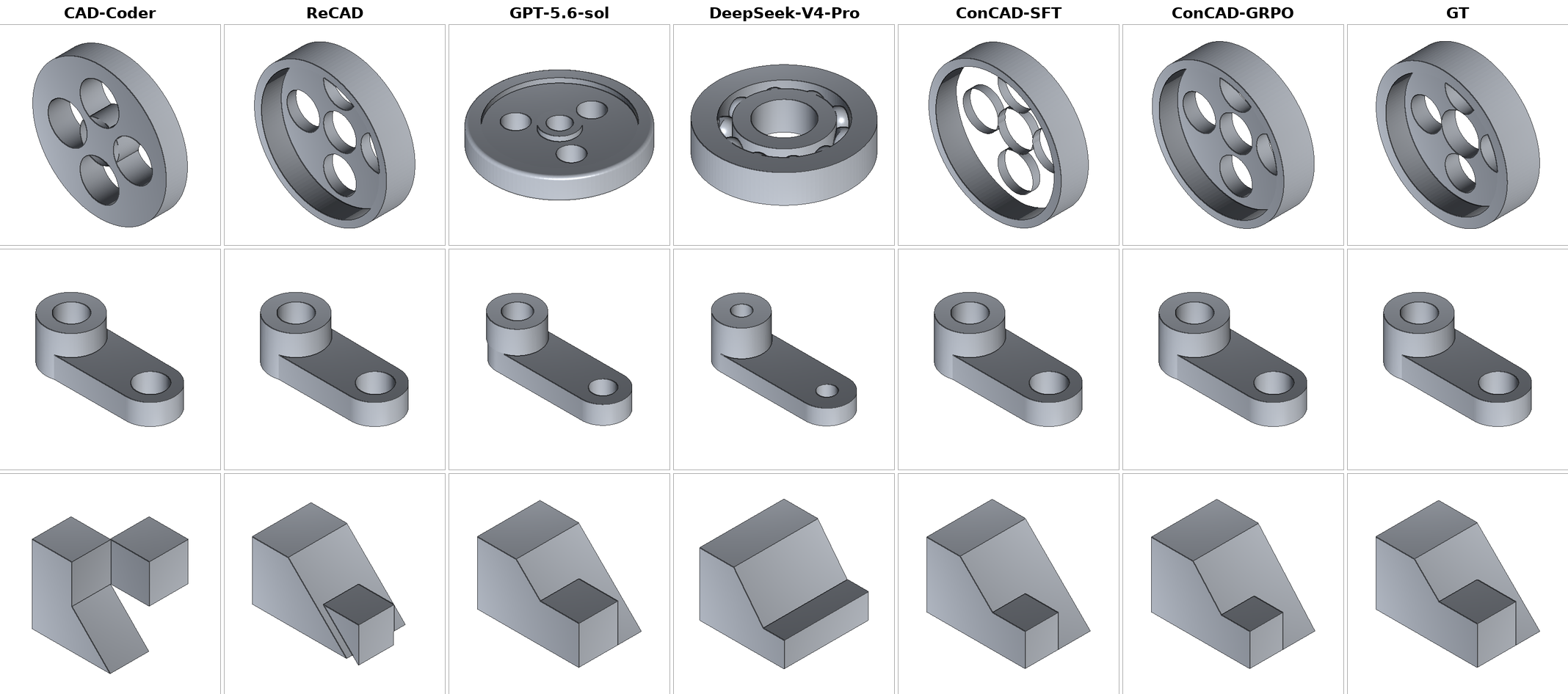} 
\caption{Reconstruction examples produced by different methods on the DeepCAD test set.}
\label{fig:qualitative}
\end{figure*}

\subsection{Main Results}

We compare ConCAD against both general foundation models and CAD-specialized baselines. During inference, all models are evaluated on a single NVIDIA H20-96GB GPU with a maximum generation limit of 4096 tokens and a sampling temperature of $T=0.3$.

Table~\ref{tab:deepcad} presents the evaluation results on the DeepCAD test set. Domain specialization proves decisive across all metrics. Without specialized training, the open-source vanilla Qwen3-VL-2B\cite{yang2025qwen3} struggles with domain-specific grammar, yielding a prohibitive invalidity ratio of 98\%. Meanwhile, although proprietary general models such as GPT-5.6-Sol\cite{singh2025openai} and DeepSeek-V4-Pro\cite{xu2026deepseek} produce mostly executable code, they lack strict parametric and geometric grounding, lagging significantly behind specialized methods in both IoU and G-CSR. In contrast, CAD-specialized architectures reliably capture CadQuery syntax while accurately preserving underlying design intent.

Compared to the supervised SFT initialization, ConCAD achieves comprehensive improvements across all primary geometric metrics, boosting IoU from 0.717 to 0.734, lowering Chamfer Distance to 0.009, and completely eliminating execution failures with an IR of 0\%. On the G-CSR metric, it also reaches the best performance at 0.913. Because this metric is not used as an optimization reward during reinforcement learning and evaluates the intrinsic B-rep topological constraints of the resulting solids, this confirms that our policy captures deep parametric design intent rather than merely fitting surface-level syntax. Relative to ReCAD, which relies on rendered image similarity, our dual-granularity reinforcement learning approach achieves superior solid fidelity and higher topological constraint satisfaction. This demonstrates that dense code-level proxy signals effectively guide policy exploration toward solid models that satisfy rigorous engineering standards. Figure~\ref{fig:qualitative} illustrates qualitative reconstructions across representative geometric structures.
\begin{table}[t]
\centering
\caption{Results on the DeepCAD test set, evaluated using the identical test split as CAD-Coder~\cite{doris2026cad}.}
\label{tab:deepcad}
\vspace{3pt}
\footnotesize
\setlength{\tabcolsep}{4.0pt}
\begin{tabular}{lcccc}
\hline
Method & IR$\downarrow$ & IoU$\uparrow$ & CD$\downarrow$ & G-CSR$\uparrow$ \\
\hline
GPT-5.6-Sol\cite{singh2025openai} & 2 & 0.599 & 0.016 & 0.879 \\
DeepSeek-V4-Pro\cite{xu2026deepseek}& 4 & 0.466 & 0.027 & 0.798 \\
Qwen3-VL-2B\cite{yang2025qwen3} & 98 & 0.160 & 0.106 & 0.850 \\
CAD-Coder~\cite{doris2026cad} & 2 & 0.662 & 0.022 & 0.837 \\
ReCAD~\cite{li2026recad} & 2 & 0.723 & 0.011 & 0.906 \\
Ours: SFT & 1 & 0.717 & 0.011 & 0.903 \\
Ours: ConCAD & \textbf{0} & \textbf{0.734} & \textbf{0.009} & \textbf{0.913} \\
\hline
\end{tabular}

% 两个表之间的垂直间距由这行完全控制，数值可自由调整：
\vspace{4pt}

\caption{Results on the Zero2CAD test set.}
\label{tab:zero2}
\vspace{3pt}
\begin{tabular}{lcccc}
\hline
Method & IR$\downarrow$ & IoU$\uparrow$ & CD$\downarrow$ & G-CSR$\uparrow$ \\
\hline
Zero2CAD~\cite{ataei2026zero} & 40 & 0.357 & 0.198 & 0.535 \\
Ours: SFT & 23 & 0.443 & 0.016 & 0.768 \\
Ours: ConCAD & \textbf{23} & \textbf{0.507} & \textbf{0.013} & \textbf{0.781} \\
\hline
\end{tabular}
\end{table}

\vspace{-8pt}
Table~\ref{tab:zero2} reports performance on the Zero2CAD-100K test set, where we evaluate on a test subset using only single-view inputs. On this benchmark, starting from the supervised SFT checkpoint, our proposed dual-granularity reinforcement learning method brings marked gains. ConCAD elevates volumetric IoU from 0.443 to 0.507, reduces Chamfer Distance to 0.013, and improves the topological constraint score G-CSR to 0.781. These results demonstrate that our approach also achieves compelling performance on more challenging Image-to-CAD tasks.

\subsection{Ablation Study}

To thoroughly investigate the contributions of individual components in ConCAD, we conduct controlled ablation experiments on the DeepCAD test set. All ablations share the identical supervised SFT initialization, curriculum filtering, rollout configuration, optimizer hyperparameters, and training budget.

Table~\ref{tab:reward-ablation} decouples the individual terms in the reward function. Applying executable volume overlap alone via $R_{\rm exec} + R_{\rm iou}$ leads to a slight performance drop compared to the SFT baseline, reflecting the sparsity of 3D spatial boolean feedback. Relying exclusively on code-level constraints via $R_{\rm exec} + R_{\rm con}$ preserves topological relations well but lacks continuous spatial grounding, causing IoU to drop to 0.654. Notably, directly employing the B-rep G-CSR metric as an online reward, denoted as $R_{\rm GCSR}$, in the formulation $R_{\rm exec} + R_{\rm GCSR} + R_{\rm iou}$ reaches an IoU of only 0.702. Because G-CSR evaluation is strictly gated by solid reconstruction, execution errors cut off structural signals, disrupting informative credit assignment. In contrast, our full formulation combining the code-level constraint reward $R_{\rm con}$ with volumetric IoU achieves the best overall performance with 0.734 IoU and 0.913 G-CSR, proving that code-level constraints serve as an effective execution-free prior.

\begin{table}[t]
\centering
\caption{Reward-component ablation on the DeepCAD test set.}
\label{tab:reward-ablation}
\vspace{2pt}
\footnotesize
\setlength{\tabcolsep}{4.0pt}
\begin{tabular}{lcccc}
\hline
Reward Formulation & IR$\downarrow$ & IoU$\uparrow$ & CD$\downarrow$ & G-CSR$\uparrow$ \\
\hline
SFT & 1 & 0.717 & 0.011 & 0.903 \\
$R_{\rm exec} + R_{\rm iou}$ & 1 & 0.700 & 0.013 & 0.897 \\
$R_{\rm exec} + R_{\rm con}$ & 1 & 0.654 & 0.010 & 0.908 \\
$R_{\rm exec} + R_{\rm GCSR} + R_{\rm iou}$ & 2 & 0.702 & 0.009 & 0.910 \\
$R_{\rm exec} + R_{\rm con} + R_{\rm iou}$ & \textbf{0} & \textbf{0.734} & \textbf{0.009} & \textbf{0.913} \\
\hline
\end{tabular}

% 控制两个表格之间的上下距离，数值可按需微调
\vspace{6pt}

\caption{Code-level constraint reward coefficient ablation on the DeepCAD test set. Setting $\lambda_c=0$ corresponds to the overlap-only formulation.}
\label{tab:lambda-ablation}
\vspace{2pt}
\begin{tabular}{ccccc}
\hline
$\lambda_c$ & IR$\downarrow$ & IoU$\uparrow$ & CD$\downarrow$ & G-CSR$\uparrow$ \\
\hline
0.0 & 1 & 0.700 & 0.013 & 0.897 \\
0.1 & 3 & 0.705 & 0.011 & \textbf{0.919} \\
0.3 & \textbf{0} & \textbf{0.734} & \textbf{0.009} & 0.913 \\
0.5 & 2 & 0.721 & 0.009 & 0.899 \\
\hline
\end{tabular}
\end{table}
% 控制合并后的整体与下方正文的间距

Table~\ref{tab:lambda-ablation} studies the sensitivity of the code-level constraint reward weighting coefficient $\lambda_c$. Balancing the reward at $\lambda_c=0.3$ establishes an optimal synergy between code-level guidance and spatial volume alignment, yielding the lowest IR, highest IoU, and robust topological fidelity. Increasing the coefficient excessively to 0.5 overemphasizes syntactic pattern matching at the expense of global shape accuracy.

\section{Conclusion}
In this paper, we presented ConCAD, a constraint-aware framework optimizing editable parametric CAD generation via GRPO. The policy is trained with dual-granularity supervision combining the code-level constraint reward and the execution-level geometry reward. Furthermore, we introduced G-CSR to rigorously quantify relational design constraints directly on B-rep topology. Evaluations across DeepCAD and Zero2CAD show that ConCAD achieves state-of-the-art performance in both geometric fidelity and topological constraint preservation, paving the way toward production-ready generative CAD modeling.
% 在本文中，我们提出了 ConCAD——一个约束感知的图像到 CAD 生成框架。该框架将图像到 CAD 的逆向建模任务构建为可编辑参数化程序的生成，并通过组相对策略优化（GRPO）进行策略学习。为了解决脆弱的内核执行信号与缺乏空间接地的代码生成之间的权衡问题，我们设计了一种双粒度奖励机制：在内核执行前从八种关系类别中提取密集结构反馈的代码级约束奖励，与通过惯量归一化保证连续三维空间保真度的执行级几何对齐奖励相协同。此外，我们提出了边界表示几何约束满足率（G-CSR），直接在 B-rep 拓扑层面上严格量化实体间的关联设计意图。在 DeepCAD 和 Zero2CAD 基准数据集上的大量实验表明，ConCAD 在严格保持工程设计约束的同时，取得了最先进的体积重合度与倒角距离（Chamfer Distance），为迈向工业级、可编辑的三维生成式 CAD 建模开辟了一条切实可行的路径。

\vfill\pagebreak

\bibliographystyle{IEEEbib}
\bibliography{refs}

\end{document}